\documentclass[
twocolumn,
]{ceurart}

\usepackage{tabularx} 
\usepackage{amsmath}  

\DeclareMathOperator*{\argmax}{arg\,max} 

\usepackage{amsfonts}
\usepackage{graphicx} 
\usepackage{subfig}

\usepackage[utf8]{inputenc} 

\begin{document}




\conference{$^1$ When citing this work, please refer to the  peer-reviewed version of this
paper, published in the Proceedings of the CIKM 2020 Conference, Vol 2699. link: \href{http://ceur-ws.org/Vol-2699/paper03.pdf}{http://ceur-ws.org/Vol-2699/paper03.pdf}}

\title{OptiLIME: Optimized LIME Explanations for Diagnostic Computer Algorithms$^1$}

\author[1,2]{Giorgio Visani}[%
orcid=0000-0001-6818-3526,
]
\ead{giorgio.visani2@unibo.it}
\address[1]{University of Bologna, School of Informatics \& Engineering, viale Risorgimento 2, 40136 Bologna (BO), Italy}
\address[2]{CRIF S.p.A., via Mario Fantin 1-3, 40131 Bologna (BO), Italy}

\author[2]{Enrico Bagli}[%
orcid=0000-0003-3913-7701,
]

\author[1]{Federico Chesani}[%
orcid=0000-0003-1664-9632,
]

\begin{abstract}
Local Interpretable Model-Agnostic Explanations (LIME) is a popular method to perform interpretability 
of any kind of Machine Learning (ML) model. It explains one ML prediction at a time, by learning a simple linear model around the prediction. The model is trained on randomly generated data points, sampled from the training dataset distribution and weighted according to the distance from the reference point - the one being explained by LIME. Feature selection is applied to keep only the most important variables, their coefficients are regarded as explanation.
LIME is widespread across different domains, although its instability - a single prediction may obtain different explanations - is one of the major shortcomings. This is due to the randomness in the sampling step, as well and determines a lack of reliability in the retrieved explanations, making LIME adoption problematic. In Medicine especially, clinical professionals trust is mandatory to determine the acceptance of an explainable algorithm, considering the importance of the decisions at stake and the related legal issues. In this paper, we highlight a trade-off between explanation's stability and adherence, namely how much it resembles the ML model. 
Exploiting our innovative discovery, we propose a framework to maximise stability, while retaining a predefined level of adherence. 
OptiLIME provides freedom to choose the best adherence-stability trade-off level and more importantly, it clearly highlights the mathematical properties of the retrieved explanation. As a result, the practitioner is provided with tools to decide whether the explanation is reliable, according to the problem at hand. We extensively test OptiLIME on a toy dataset - to present visually the geometrical findings - and a medical dataset. In the latter, we show how the method comes up with meaningful explanations both from a medical and mathematical standpoint.
\end{abstract}


\maketitle

\section{Introduction}

Nowadays Machine Learning (ML) is pervasive and widespread across multiple domains. Medicine makes no difference, on the contrary it is considered one of the greatest challenges of Artificial Intelligence \citep{holzinger_causability_2019}. The idea of exploiting computers to provide assistance to the medical personnel is not new. An historical overview on the topic, starting from the early `60s is provided in \cite{kononenko_machine_2001}. More recently, computer algorithms have been proven useful for patients and medical concepts representation \cite{miotto_deep_2016}, outcome prediction \cite{choi_doctor_2016},\cite{rajkomar_scalable_2018},\cite{shickel_deep_2017} and new phenotype discovery \cite{che_deep_2015},\cite{lasko_computational_2013}. An accurate overview of ML successes in Health related environments, is provided by Topol in \cite{topol_high-performance_2019}.
\par

Unfortunately, ML methods are hardly perfect and, especially in the medical field where human lives are at stake, Explainable Artificial Intelligence (XAI) is urgently needed \cite{holzinger_machine_2018-1}.
Medical education, research and accountability (``who is accountable for wrong decisions?'') are some of the main topics XAI tries to address. To achieve the explainability, quite a few techniques have been proposed in recent literature. These approaches can be grouped based on different criterion \cite{molnar_interpretable_2020}, \cite{guidotti_survey_2018} such as i) Model
agnostic or model specific ii) Local, global or example based iii) Intrinsic or post-hoc iv) Perturbation or saliency based.
Among them, model agnostic approaches are quite popular in practice, since the algorithm is designed to be effective on any type of ML model.
\par

LIME \cite{ribeiro_why_2016} is a well-known instance-based, model agnostic algorithm.
The method generates data points, sampled from the training dataset distribution and weighted according to distance from the instance being explained. Feature selection is applied to keep only the most important variables and a linear model is trained on the weighted dataset. The model coefficients are regarded as explanation. LIME has already been employed several times in medicine, such as on Intensive Care data \cite{katuwal_machine_2016} and cancer data \cite{zhang_development_2018},\cite{moreira_investigation_2020}. The technique is known to suffer from instability, mainly caused by the randomness introduced in the sampling step. Stability is a desirable property for an interpretable model, whereas the lack of it reduces the trust in the explanations retrieved, especially in the medical field.
\par

In our contribution, we review the geometrical idea on which LIME is based upon. Relying on statistical theory and simulations, we highlight a trade-off between the explanation's stability and adherence, namely how much LIME's simple model resembles the ML model. Exploiting our innovative discovery, we propose OptiLIME: a framework to maximise the stability, while retaining a predefined level of adherence. OptiLIME provides both i) freedom to choose the best adherence-stability trade-off level and ii) it clearly highlights the mathematical properties of the explanation retrieved. 
As a result, the practitioner is provided with tools to decide whether each explanation is reliable, according to the problem at hand.
\par

We test the validity of the framework on a medical dataset, where the method comes up with meaningful explanations both from a medical and mathematical standpoint. In addition, a toy dataset is employed to present visually the geometrical findings.
\par
The code used for the experiments is available at \\ 
\href{https://github.com/giorgiovisani/LIME_stability}{https://github.com/giorgiovisani/LIME\_stability}.

\section{Related Work}

For the sake of shortness, in the following review  we consider only model agnostic techniques, which are effective on any kind of ML model by construction.
A popular approach is to exclude a certain feature, or group of features, from the model and evaluate the loss incurred in terms of model goodness. Such value quantifies the importance of the excluded feature: an high loss value underlines an important variable for the prediction task. The idea has been first introduced by Breiman \cite{breiman_random_2001} for the Random Forest model and has been generalised to a model-agnostic framework, named LOCO \cite{lei_distribution-free_2018}. Based on variable exclusion, the predictive power of the ML models has been decomposed into single variables contribution in PDP \cite{friedman_greedy_2001}, ICE \cite{goldstein_peeking_2015} and ALE \cite{apley_visualizing_2016} plots, based on different assumptions about the ML model. The same idea is exploited also for local explanations in SHAP \cite{lundberg_unified_2017-1}, where the decomposition is obtained through a game-based setting.
\par
Another common approach is to train a surrogate model mimicking the behaviour of the ML model. In this vein, approximations on the entire input space are provided in \cite{craven_extracting_1996} and \cite{zhou_interpreting_2016-1} among others, while LIME \cite{ribeiro_why_2016} and its extension using decision rules \cite{ribeiro_anchors_2018} rely on this technique for providing local approximations.

\subsection{LIME Framework}

A thorough examination of LIME is provided from a geometrical perspective, while a detailed algorithmic description can be found in \cite{ribeiro_why_2016}. We may consider the ML model as a multivariate surface in the $\mathbb{R}^{d+1}$ space spanned by the $d$ independent variables $X_1,...,X_d$ and the $Y$ dependent variable. \par
LIME's objective is to find the tangent plane to the ML surface, in the point we want to explain. This task is analytically unfeasible, since we don't have a parametric formulation of the function, besides the ML surface may have a huge number of discontinuity points, preventing the existence of a proper derivative and tangent.
To find an approximation of the tangent, LIME uses a Ridge Linear Model to fit points on the ML surface, in the neighbourhood of the reference individual.
\par
Points all over the $\mathbb{R}^{d}$ space are generated, sampling the $\mathbf{X}$ values from a Normal distribution inferred from the training set. The $Y$ coordinate values are obtained by ML predictions, so that the generated points are guaranteed to perfectly lie on the ML surface. The concept of neighbourhood is introduced using a kernel function (RBF Kernel), which smoothly assigns higher weights to points closer to the reference. Ridge Model is trained on the generated dataset, each point weighted by the kernel function, to estimate the linear relationship $\mathbf{E}(Y) = \alpha + \sum_{j=1}^d \beta_j X_j$. The $\beta$ coefficients are regarded as LIME explanation.
\par

\subsection{LIME Instability}
One of the main issues of LIME is the lack of stability.\\
Explanations derived from repeated LIME calls, under the same conditions, are considered stable when statistically equal \cite{visani_statistical_2020-1}. In \cite{alvarez-melis_robustness_2018-1} the authors provide insight about LIME's lack of robustness, a similar notion to the above-mentioned stability. Analogous findings also in \cite{gosiewska_ibreakdown_2019}.
Often, practitioners are either not aware of such drawback or diffident about the method because of its unreliability. By all means, unambiguous explanations are a key desiderata for the interpretable frameworks.
\par

The major source of LIME instability comes from the sampling step, when new observations are randomly selected. Some approaches, grouped in two high level concepts, have been recently laid out in order to solve the stability issue.

\subsubsection*{Avoid the sampling step}

In \cite{zafar_dlime_2019} the authors propose to bypass the sampling step using the training units only and a combination of Hierarchical Clustering and K-Nearest Neighbour techniques. Although this method achieves stability, it may find a bad approximation of the ML function, in regions with only few training points.

\subsubsection*{Evaluate the post-hoc stability}

The shared idea is to repeat LIME method at the same conditions, and test whether the results are equivalent. Among the various propositions on how to conduct the test, in \cite{shankaranarayana_alime_2019-1} the authors compare the standard deviations of the Ridge coefficients, whereas \cite{molnar_limitations_2020} examines the stability of the feature selection step - whether the selected variables are the same - . In \cite{visani_statistical_2020-1} two complementary indices have been developed, based on statistical comparison of the Ridge models generated by repeated LIME calls. The Variables Stability Index (VSI) checks the stability of the feature selection step, whereas the Coefficients Stability Index (CSI) asserts the equality of coefficients attributed to the same feature.

\section{Methodology}

OptiLIME consists in a framework to guarantee the highest reachable level of stability, constrained to the finding of a relevant local explanation.
From a geometrical perspective, the relevance of the explanation corresponds to the adherence of the linear plane to the ML surface. To evaluate the stability we rely on the CSI and VSI indices \cite{visani_statistical_2020-1}, while the adherence is assessed using the $R^2$ statistic, which measures the goodness of the linear approximation through a set of points \cite{greene_econometric_2003}. All the figures of merit above span in the range $[0,1]$, where higher values define respectively higher stability and adherence.
\par
To fully explain the rationale of the proposition, we first cover three important concepts about LIME.
In this section we employ a Toy Dataset to show our theoretical findings.

\subsection*{Toy Dataset}

\begin{figure}
\centering
\includegraphics[width=\columnwidth]{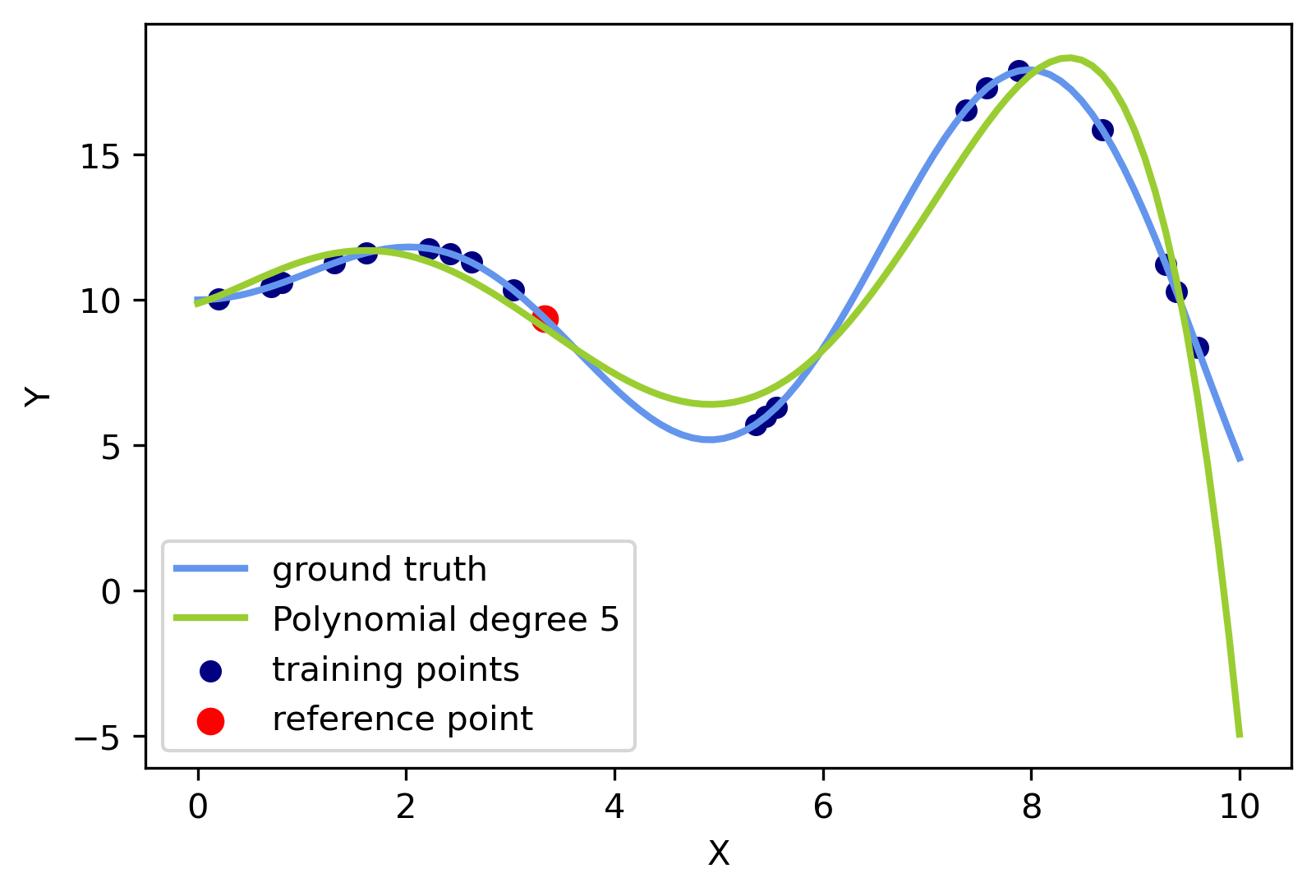}
\caption{Toy Dataset}
\label{Toy_Dataset}
\end{figure}

The dataset is generated from the Data Generating Process:
\begin{equation*}
 	 Y = sin(X)*X + 10
 \end{equation*}
	100 distinct points have been generated uniformly in the $X$ range [0,10] and only 20 of them were kept, at random. In Figure \ref{Toy_Dataset}, the blue line represents the True DGP function, whereas the green one is its best approximation using a Polynomial Regression of degree 5 on the generated dataset (blue points). In the following we will regard the Polynomial as our ML function, we will not make use of the True DGP function (blue line) which is usually not available in practical data mining scenarios. The red dot is the reference point in which we will evaluate the local LIME explanation. The dataset is intentionally one dimensional, so that the geometrical ideas about LIME may be well represented in a 2d plot.

\subsection{Kernel Width defines locality}
Locality is enforced through a kernel function, the default is the RBF Kernel (Formula \ref{eq1}). It is applied to each point $x^{(i)}$ generated in the sampling step, obtaining an individual weight. The formulation provides smooth weights in the range $[0,1]$ and flexibility through the kernel width parameter $kw$.
\begin{equation}\label{eq1}
	RBF(x^{(i)}) = \exp \left( - \frac{||x^{(i)}-x^{(ref)}||^2}{kw} \right)
\end{equation}
The RBF flexibility makes it suitable to each situation, although it requires a proper tuning:
setting a high $kw$ value will result in considering a neighbourhood of large dimension, shrinking $kw$ we shrink the width of the neighbourhood.
\par

\begin{figure}
\centering
\includegraphics[width=\columnwidth]{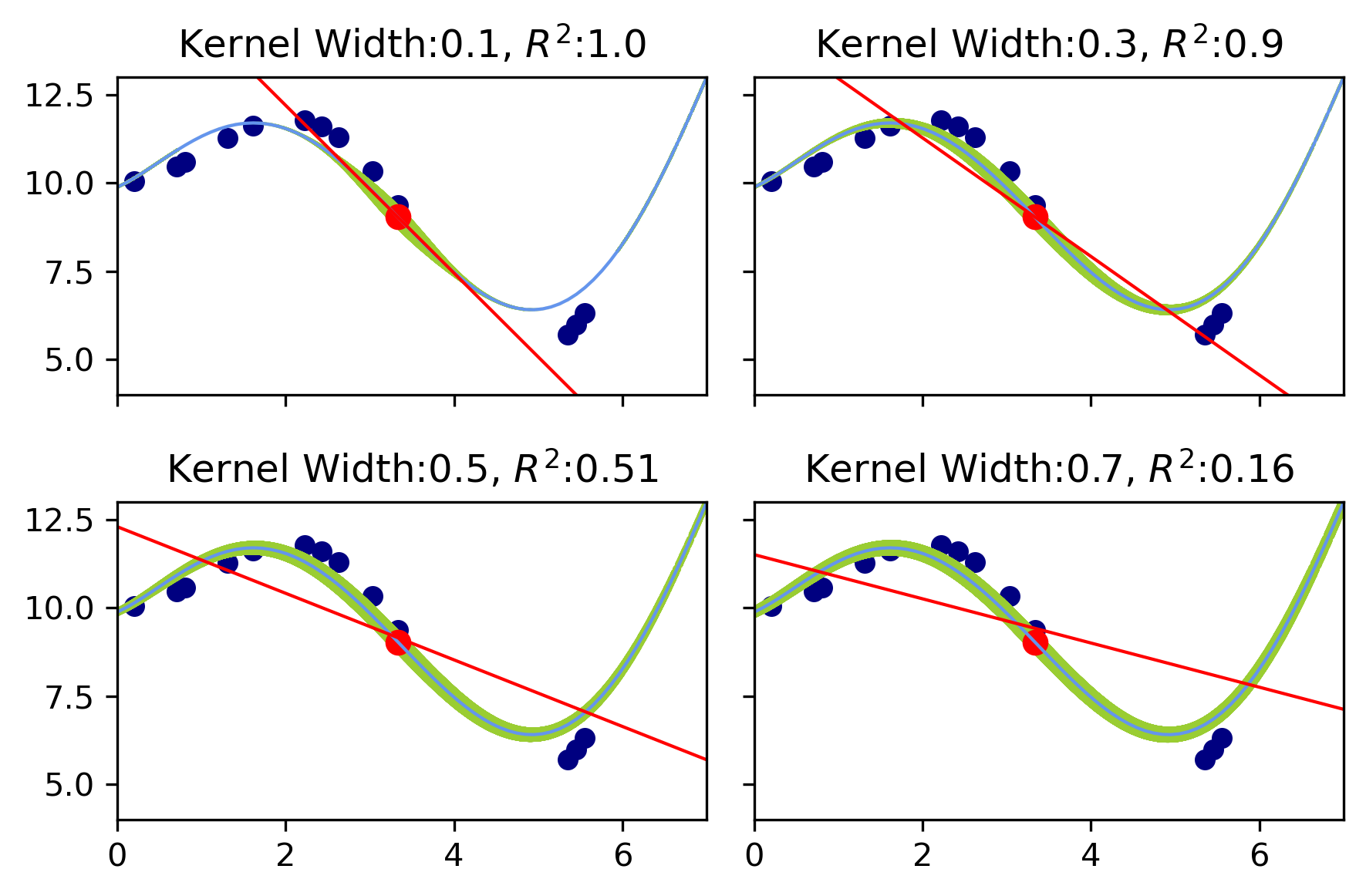}
\caption{LIME explanations for different kernel widths}
\label{Toy_KW}
\end{figure}

In Figure \ref{Toy_KW}, LIME generated points are displayed as green dots and the corresponding LIME explanations (red lines) are shown. The points are scattered all over the ML function, however their size is proportional to the weight assigned by the RBF kernel. Small kernel widths assign significant weights only to the closest points, making the further ones almost invisible. In this way, they do not contribute to the local linear model.
\par
The concept of locality is crucial to LIME: a neighbourhood too large may cause the LIME model not to be adherent to the ML function in the considered neighbourhood.

\subsection{Ridge penalty is harmful to LIME}
In statistics, data are assumed to be generated from a Data Generating Process (DGP) combined with a source of white noise, so that the standard formulation of the problem is $Y = f(\mathbf{X}) + \mathcal{E}$, where $\mathcal{E} \sim N(0,\sigma^2)$. The aim of each statistical model is to retrieve the best specification of the DGP function $f(\mathbf{X})$, given the noisy dataset.
\par
Ridge Regression \cite{hoerl_ridge_1970} assumes a linear DGP, namely $f(\mathbf{X}) = \alpha + \sum_{j=1}^d \beta_jX_j$, and applies a penalty proportional to  the norm of the $\beta$ coefficients, enforced during the estimation process through the penalty parameter $\lambda$.
This technique is useful when dealing with very noisy datasets (where the stochastic component $\mathcal{E}$ exhibits high variance $\sigma^2$) \cite{van_wieringen_lecture_2019}. In fact, the noise makes various sets of coefficients as viable solutions. Instead, tuning $\lambda$ to its proper  value allows Ridge to retrieve a unique solution.
\par

\begin{figure}
  \centering
  \subfloat[Ridge Penalty $ = 0$]{\includegraphics[width=\columnwidth]{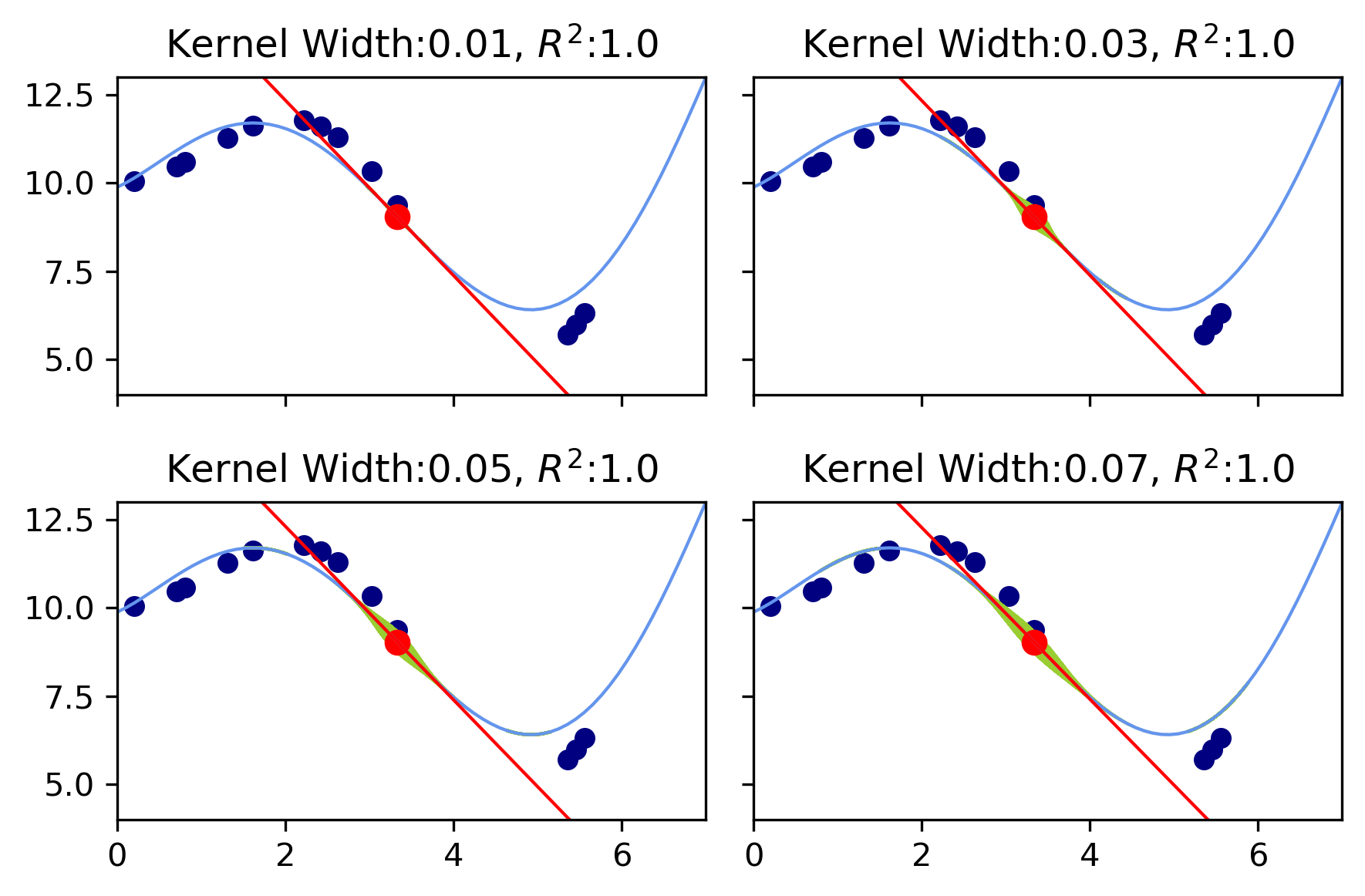}\label{ridge0}}
  \hfill
  \subfloat[Ridge Penalty $ = 1$]{\includegraphics[width=\columnwidth]{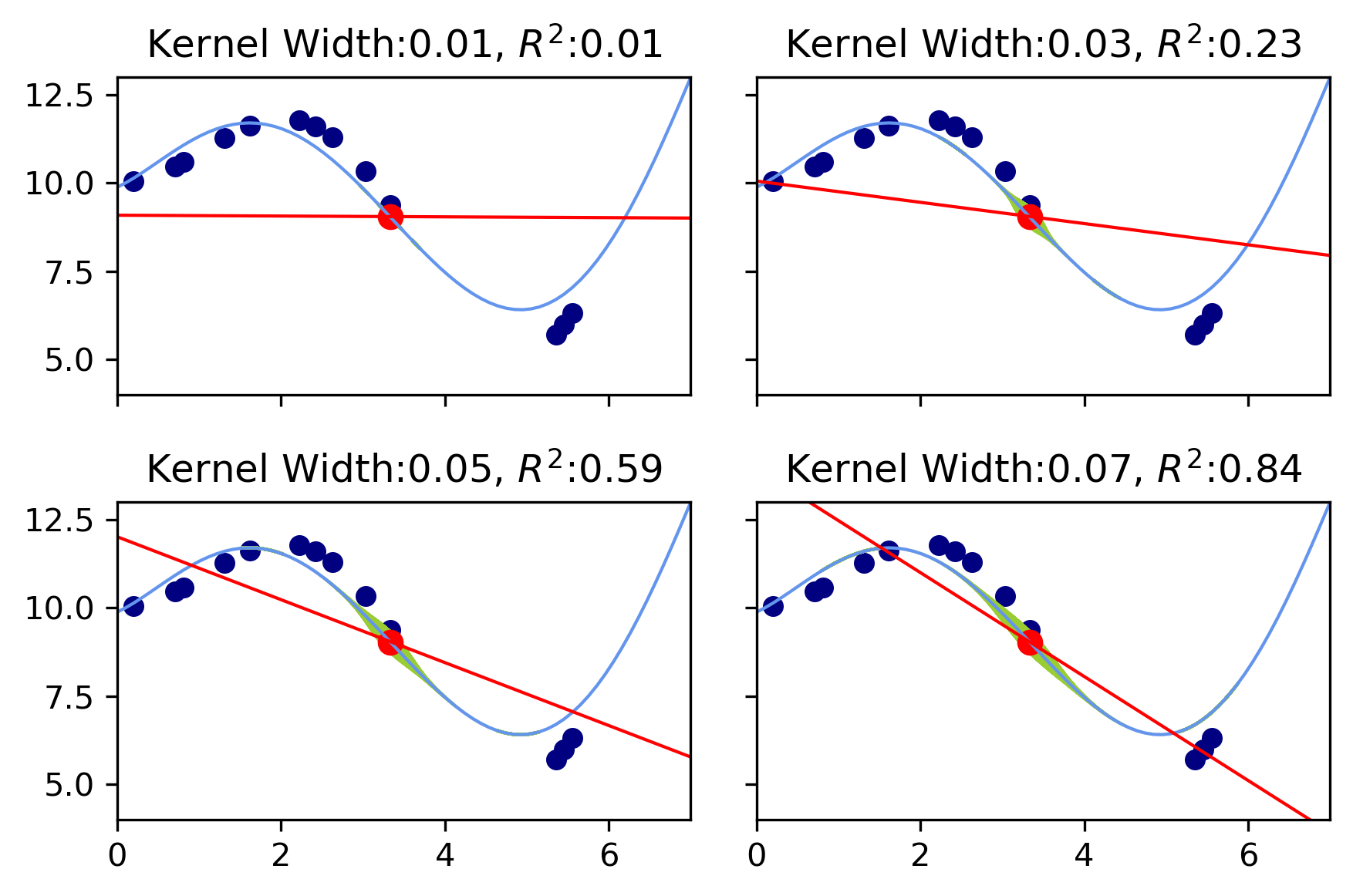}\label{ridge1}}
  \caption{Effects of Ridge Penalty on LIME explanations}
  \label{ridge_problems}
\end{figure}

In the LIME setting, the ML function acts as the DGP, while the sampled points are the dataset. Recalling that the $Y$ coordinate of each point is given by ML prediction, it is guaranteed they lie exactly on the ML surface by construction. Hence, no noise is present in our dataset. For this reason, we argue that Ridge penalty is not needed, on the contrary it can be harmful and distort the right estimates of the parameters, as shown in Figure \ref{ridge_problems}.
\par

In the \ref{ridge1} panel, Ridge penalty $\lambda=1$  (LIME default) is employed, whereas in \ref{ridge0} no penalty ($\lambda=0$) is imposed. It is possible to see how the estimation gets severely distorted by the penalty, proven also by the $R^2$ values. This happens especially for small kernel width values, since each unit has very small weight and the weighted residuals are almost irrelevant in the Ridge loss, which is dominated by the penalty term. To minimize the penalty term the coefficients are shrunk towards 0.

\subsection{Relationship between Stability, Adherence and Kernel Width}

Since the kernel width represents the main hyper-parameter of LIME, we wish to understand how Stability and Adherence vary wrt to it. \\
From the theory, we have few helpful results:
\begin{itemize}
 	\item Taylor Theorem \cite{greene_econometric_2003} gives a polynomial approximation for any differentiable function, calculated in a given point. If we truncate the formula to the first degree polynomial, we obtain a linear function, its approximation error depends on the distance from the point in which the error is evaluated and the given point. \\
 	Thus, if we assume the ML function to be differentiable in the neighbourhood of $x^{(ref)}$, the adherence of the linear model is expected to be inversely proportional to the width of the neighbourhood, i.e. to the kernel width. This is true since the approximation error depends on the distance from the two points, namely the neighbourhood size.

 	\item in Linear Regression, the standard deviation of the coefficients is inversely correlated to the standard deviation of the $\mathbf{X}$ variables \cite{greene_econometric_2003}. \\
 	The stability of the explanations depends on the spread of the $\mathbf{X}$ variables in our weighted dataset. We then expect the kernel width and Stability to be directly proportional.
 \end{itemize}

\begin{figure}
\centering
\includegraphics[width=\columnwidth]{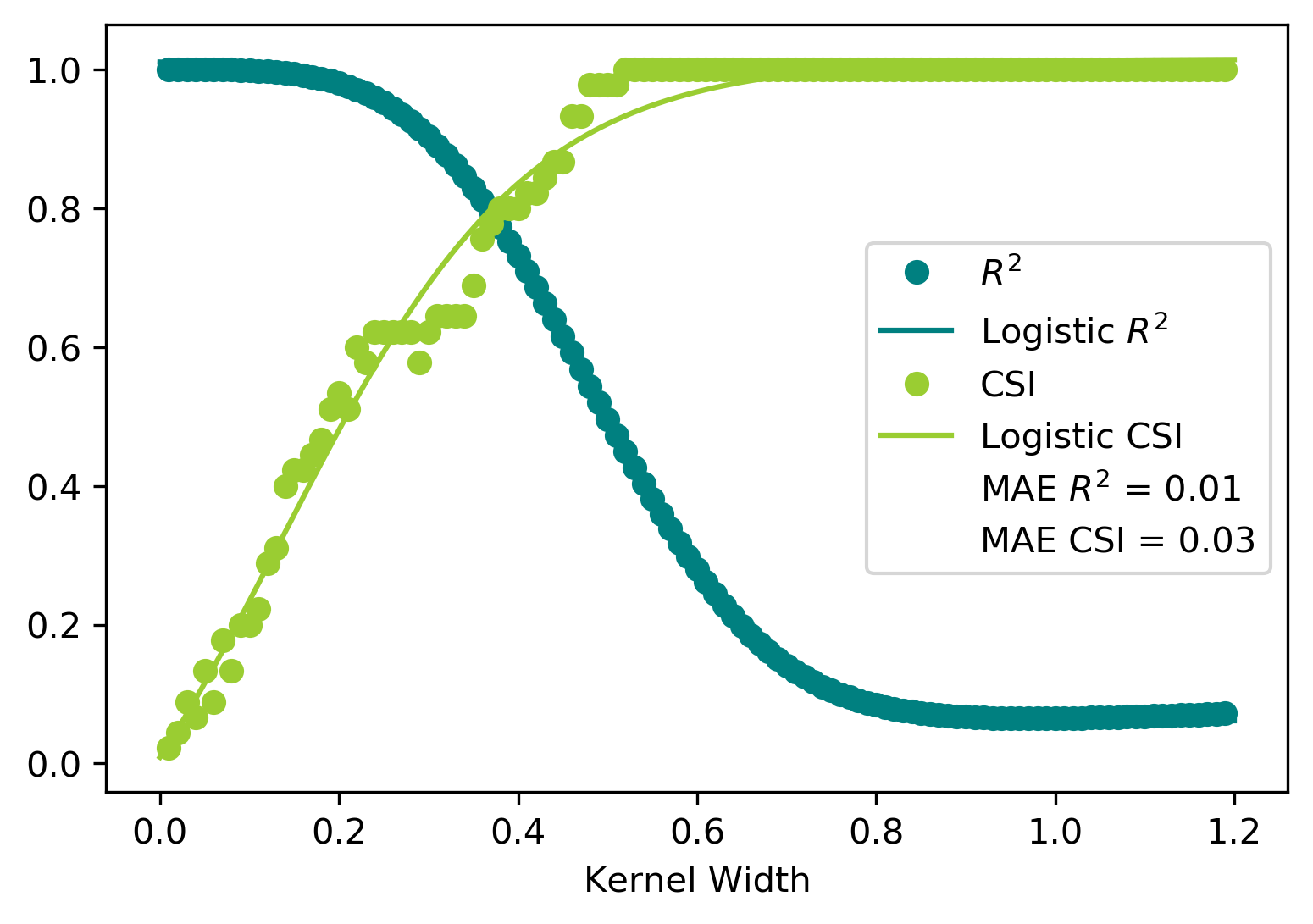}
\caption{Relationship among kernel width, $R^2$ and CSI}
\label{Monotonical_Relationship}
\end{figure}

To illustrate the conjectures above, we run LIME for different kernel width values and evaluate both $R^2$ and CSI metrics (VSI is not considered in the Toy Dataset, since only one variable is present). In Figure \ref{Monotonical_Relationship} the results of such experiment, for the reference unit, are shown.

Both the adherence and stability are noisy functions of the kernel width: they contain some stochasticity, due to the different datasets generated by each LIME call. Despite this, it is possible to detect a clear pattern: monotonically increasing for the CSI Index and monotonically decreasing for the $R^2$ statistic.
\par
For numerical evidence of these properties, we fit the Logistic function \cite{verhulst_correspondance_1838}, which retrieves the best monotonous approximation to a set of points. The goodness of the logistic approximation is confirmed by a low value of the Mean Absolute Error (MAE).\\
To corroborate our assumption, the same process has been repeated on all the units of the Toy Dataset, obtaining average MAE for the $R^2$ approximation of 0.005 and for the CSI of 0.026. The logistic growth rate has also been inspected: $R^2$ highest growth rate is -10.78 and CSI lowest growth rate is 7.20. These results ensure the monotonous relationships of adherence and stability with the kernel width, respectively decreasing and increasing.

\subsection{OptiLIME}

Previously, we empirically showed that adherence and stability are monotonous noisy functions of the kernel width: for increasing kernel width we observe, on average, decreasing adherence and increasing stability.
\par

Our proposition consists in a framework which enables the best choice for the trade-off between stability and adherence of the explanations. OptiLIME sets a desired level of adherence and finds the largest kernel width, matching the request. At the same time, the best kernel width provides the highest stability value, constrained to the chosen level of adherence.
At the end of the day, OptiLIME consists in an automated way of finding the best kernel width. Moreover, it empowers the practitioner to be in control of the trade-off between the two most important properties of LIME Local Explanations.
\par

To retrieve the best width, OptiLIME converts the decreasing $R^2$ function into $l(kw,\tilde{R}^2)$, by means of Formula \ref{loss_function}:

\begin{equation}\label{loss_function}
	l(kw,\tilde{R}^2)=
\begin{cases}
R^2(kw), \qquad &\text{if}\; R^2(kw) \le \tilde{R}^2 \\
2\tilde{R}^2 - R^2(kw) & \text{if}\; R^2(kw) > \tilde{R}^2
\end{cases}
\end{equation}
where $\tilde{R}^2$ is the requested adherence.\\
For a fixed $\tilde{R}^2$, chosen by the practitioner, the function $l(kw,\tilde{R}^2)$ presents a global maximum. We are particularly interested in the $\argmax_{kw} l(kw,\tilde{R}^2)$, namely the best kernel width.
\par
In order to solve the optimum problem, Bayesian Optimization is employed, since it is the most suitable technique to find the global optimum of noisy functions \cite{letham_constrained_2019}. The technique relies on two parameters to be set beforehand: $p$, number of preliminary calls with random $kw$ values, $m$, number of iterations of the search refinement strategy. Increasing the parameters ensures to find a better kernel width value, at the cost of longer computation time.\\

\begin{figure}
\centering
\includegraphics[width=\columnwidth]{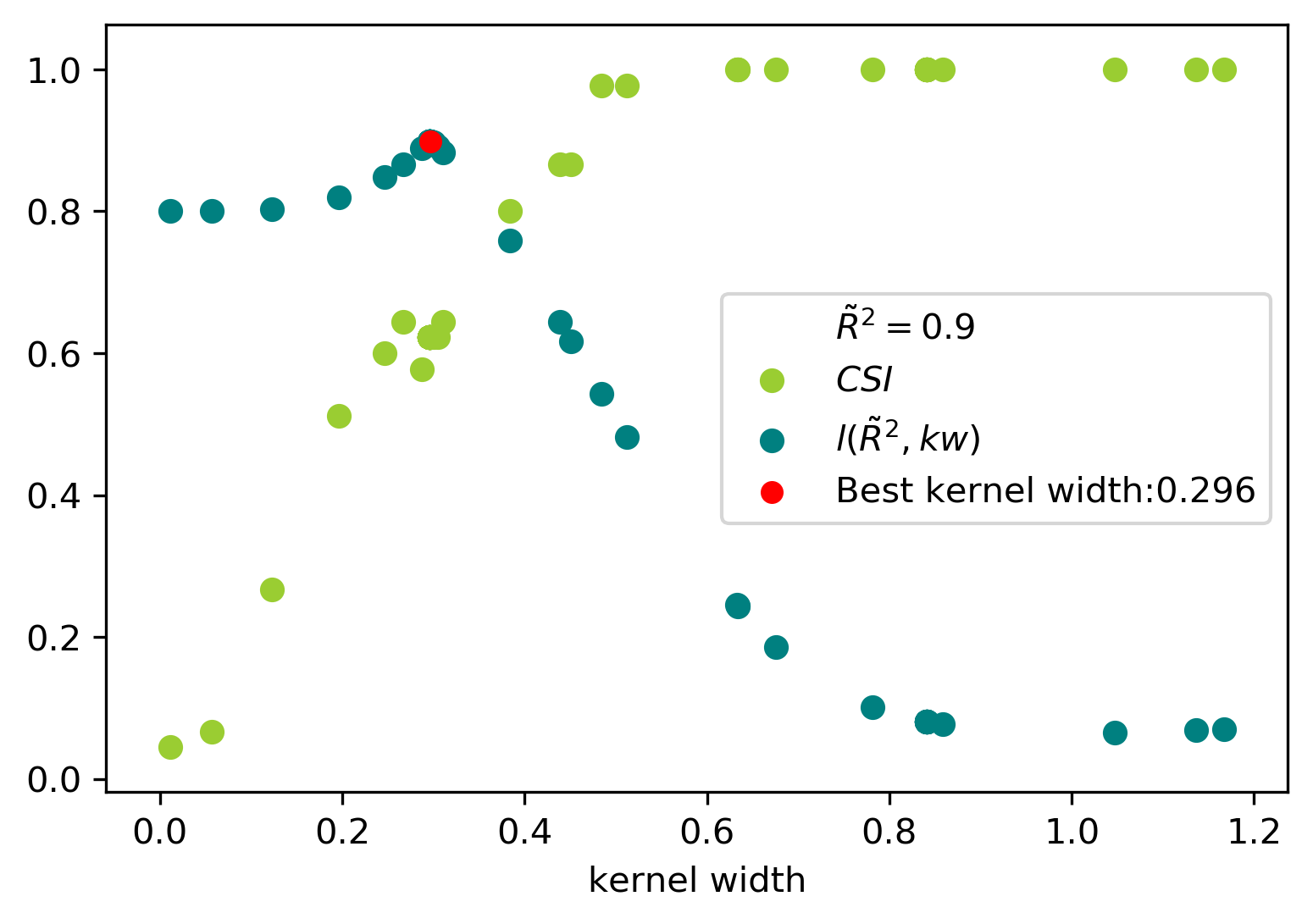}
\caption{OptiLIME Search for the best kernel width}
\label{Bayes_opt}
\end{figure}

In Figure \ref{Bayes_opt}, an application of OptiLIME to the reference unit of the Toy Dataset is presented. $\tilde{R}^2$ has been set to 0.9, $p=20$ and $m=40$. The points in the plot represent the distinct evaluations performed by the Bayesian Search in order to find the optimum. \\
Comparing the plot with Figure \ref{Monotonical_Relationship}, we observe the effect of Formula \ref{loss_function} on the left part of the $R^2$ and $l(kw,\tilde{R}^2)$ functions. In Figure \ref{Bayes_opt} the search has converged to the maximum, evaluating various points close to the best kernel width. At the same time, it is evident the stochastic nature of the CSI function: the several CSI measurements, performed in the proximity of 0.3 value of the kernel width, show a certain variation. Nonetheless, it is possible to recall the increasing CSI trend.

\section{Case Study}

\subsection*{Dataset}
To validate our methodology we use a well known medical dataset: NHANES I. It has been employed for medical research  \cite{fang_serum_2000},\cite{launer_body_1994} as well as a benchmark to test explanation methods \cite{lundberg_local_2020}.
The original dataset is described in \cite{cox_plan_1992-1}. We use a reformatted version, released at \href{https://github.com/suinleelab/treeexplainer-study}{http://github.com/suinleelab/treexplainer-study}. It contains 79 features, based on clinical measurements of 14,407 individuals. The aim is to model the risk of death over twenty years of follow-up.

\subsection*{Diagnostic Algorithm}
Following Lundberg \cite{lundberg_local_2020} prescriptions, the dataset has been divided into a 64/16/20 split for
train/validation/test. The features have been mean imputed and standardized based on statistics computed on the training set. A Survival Gradient Boosting model has been trained, using the XGBoost framework \cite{chen_xgboost_2016-1}. Its hyper-parameters have been optimized by coordinate descent, using the C-statistic \cite{heagerty_timedependent_2000} on the validation set as the figure of merit.

\subsection*{Explanations}
We use the OptiLIME framework to achieve the optimal explanation of the XGBoost model on the dataset. We consider two randomly chosen individuals to visually show the results.
In our simulation, we consider 0.9 as a reasonable level of adherence. OptiLIME is employed to find the proper kernel width to achieve $R^2$ value close to $0.9$ while maximizing stability indices for the local explanation models.

\begin{figure}

\subfloat[Best LIME Explanation, Unit 100]{%
  \includegraphics[clip,width=\columnwidth]{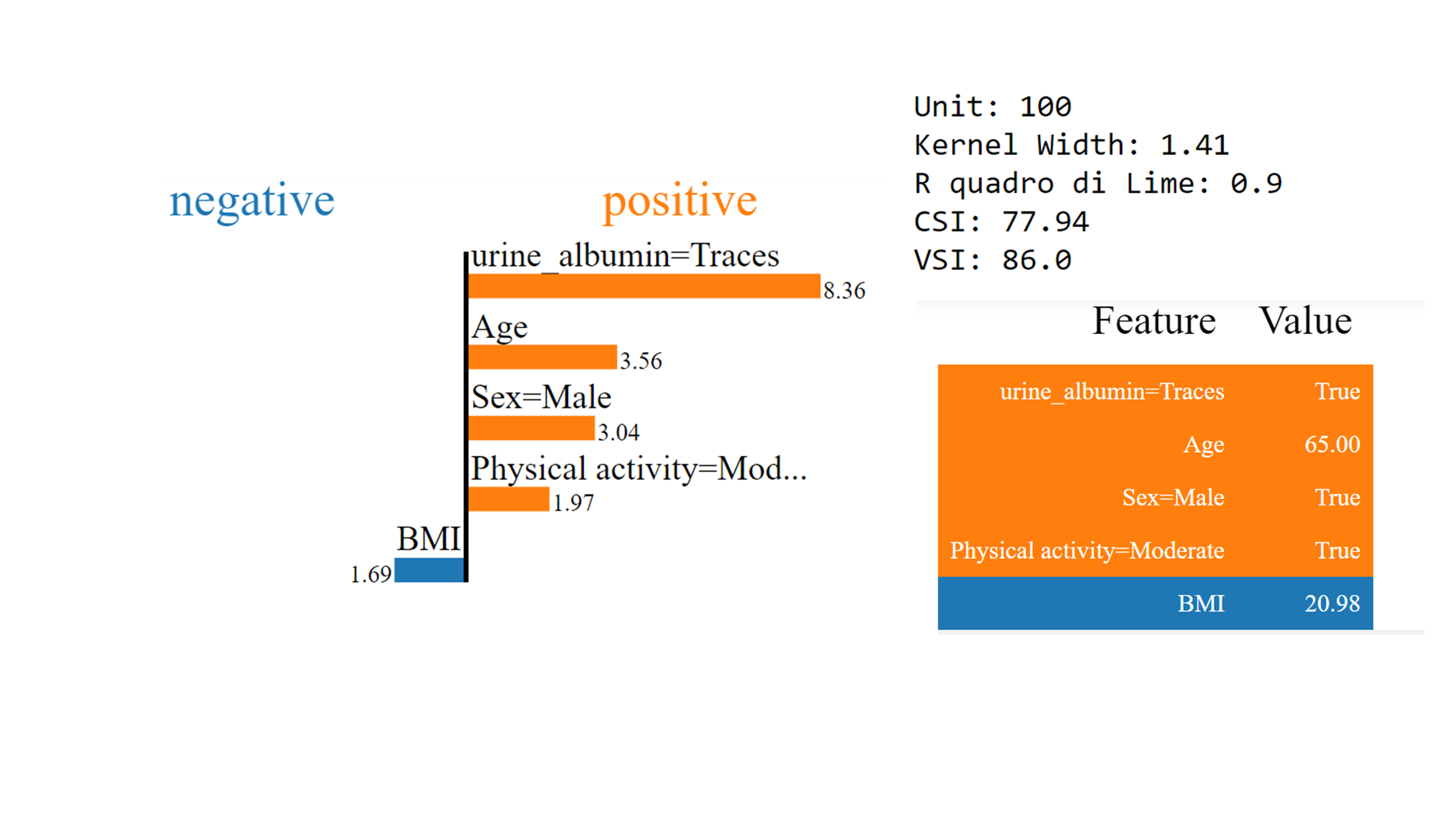}\label{LIME_unit100}%
}

\subfloat[Best LIME Explanation, Unit 7207]{%
  \includegraphics[clip,width=\columnwidth]{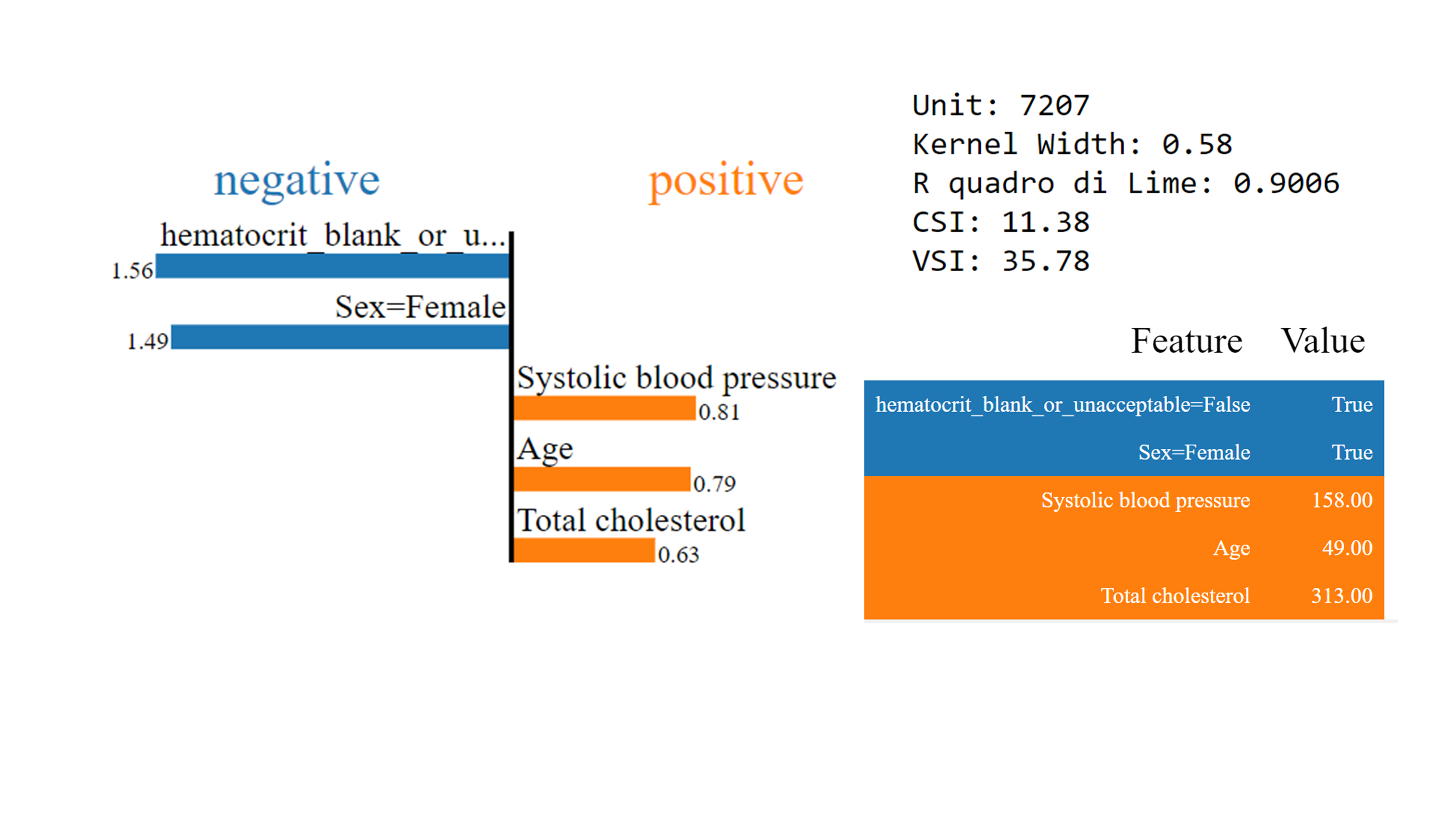}\label{LIME_unit7207}%
}{}
\caption{NHANES individual Explanations using OptiLIME}
\label{LIME_Explanations}
\end{figure}

The model prediction consists in the hazard ratio for each individual, higher prediction means the individual is likely to survive a shorter time. Therefore, positive coefficients define risk factors, whereas protective factors have negative values.
\par
LIME model interpretation is the same as a Linear Regression model, but with the additional concept of locality. As an example, for Age variable we distinguish different impact based on the individual characteristics: having 1 year more for the Unit 100 (increasing from 65 to 66 years) will raise the death risk of 3.56 base points, for Unit 7207 1 year of ageing (from 49 to 50) will increase the risk of just 0.79. Another example is the impact of Sex: it is more pronounced in elder people (being female is a protective factor for 1.49 points at age 49, at age 65 being male has a much worse impact, as a risk factor for 3.04).
\par
For the Unit 100 in Figure \ref{LIME_unit100}, the optimal kernel width is a bit higher compared with Unit 7207 in Figure \ref{LIME_unit7207}. This is probably caused by the ML model having a higher degree of non linearity for the latter unit: to achieve the same adherence, we are forced to consider a smaller portion of the ML model, hence a small neighbourhood.
Smaller kernel width implies also a reduced Stability, testified by small values of the VSI and CSI indices.
Whenever the practitioner desires more stable results, it is possible to re-run OptiLIME with a less strict requirement for the adherence. It is important to remark that low degrees of adherence will make the explanations increasingly more global: the linear surface retrieved by LIME will consist in an average of many local non-linearities of the ML model.
\par

The computation time largely depends on the Bayesian Search, controlled by the parameters $p$ and $m$. In our setting, $p=10$ and $m=30$ produce good results for both the units in Figure \ref{LIME_Explanations}. \\
On a 4 Intel-i7 CPUs 2.50GHz laptop, the OptiLIME evaluation for Unit 100 and Unit 7207 took respectively 123 and 147 seconds to compute. For faster, but less accurate results, the Bayesian Search parameters can be reduced.

\section{Conclusions}

In Medicine, diagnostic computer algorithms providing accurate predictions have countless benefits, notably they may help in saving lives as well as reducing medical costs. 
However, precisely because of the importance of these matters, the rationale of the decisions must be clear and understandable. A plethora of techniques to explain the ML decisions has grown in recent years, though there is no consensus on the best in class, since each method presents some drawbacks. Explainable models are required to be reliable, thus stability is regarded as a key desiderata.
\par

We consider the LIME technique, whose major drawback lies in the lack of stability. Moreover, it is difficult to tune properly its main parameter: different values of the kernel width provide substantially different explanations.
\par


The main contribution of this paper consists in the clear decomposition of the LIME framework in its relevant components and the exhaustive analysis of each one, starting from the geometrical meaning through the empirical experiments to validate our intuitions. We showed that Ridge penalty is not needed and LIME works best with simple Linear Regression as explainable model. In addition, smaller kernel width values provide a more adherent LIME plane to the ML surface, therefore a more realistic local explanation. Eventually, the trade-off between the adherence and stability properties is extremely valuable since it empowers the practitioner to choose the best kernel width consciously.
\par 

We exploit these findings in order to tackle LIME weak points. The result is the OptiLIME framework, which represents a new and innovative contribution to the scientific community. OptiLIME achieves stability of the explanations and automatically finds the proper kernel width value, according to the practitioner's needs.
\par

The framework may serve as an extremely useful tool: using OptiLIME, the practitioner knows how much to trust the explanations, based on their stability and adherence values.
\par
Nonetheless, we acknowledge that the optimization framework may be improved to allow for a faster and more precise computation.

\begin{acknowledgments}
We acknowledge financial support by CRIF S.p.A. and Università degli Studi di Bologna.
\end{acknowledgments}

\bibliography{Remote}


\end{document}